\documentclass[12pt]{iopart}
\expandafter\let\csname equation*\endcsname\relax
\expandafter\let\csname endequation*\endcsname\relax
\usepackage[cmex10]{amsmath}
\usepackage[font=footnotesize]{subfig}
\usepackage{amssymb}
\usepackage{graphicx}
\usepackage{fixltx2e}
\usepackage[abbr]{harvard}
\usepackage[export]{adjustbox}
\usepackage{textcomp}

\usepackage{xcolor}
\usepackage{appendix}
\usepackage{hhline}
\usepackage[ruled,lined]{algorithm2e}

\newcommand{\Tbf}{\mathbf{T}}
\newcommand{\Abf}{\mathbf{A}}
\newcommand{\Bbf}{\mathbf{B}}

\newcommand{\Cbf}{\mathbf{C}}
\newcommand{\nbf}{\mathbf{n}}
\newcommand{\Ecal}{\mathcal{E}}

\newcommand{\Ccal}{\mathcal{C}}
\newcommand{\Dcal}{\mathcal{D}}
\newcommand{\Scal}{\mathcal{S}}
\newcommand{\TRE}{\mathrm{TRE}}

\begin{document}
\title[Adaptive Local BC to improve DIR]{Adaptive local boundary conditions to improve Deformable Image Registration}

%\author[1,2]{E Inacio} 
%\author[2]{L Lafitte}
%\author[1]{L Facq}
%\author[1,2]{C Poignard}
%\author[1,2]{B Denis~de~Senneville}

%\address[1]{University of Bordeaux, IMB, UMR CNRS 5251, INRIA Project team Monc, Talence, France, F-33405 Talence Cedex, France}
%\address[2]{INRIA Bordeaux Sud Ouest, 200 Av. de la Tour, Talence, 33405, France}

\author{Eloïse Inacio$^{1,2}$, Luc Lafitte$^{2}$, Laurent Facq$^{1}$, Clair Poignard$^{1,2}$, Baudouin Denis de Senneville$^{1,2}$}
\address{$^1$ University of Bordeaux, IMB, UMR CNRS 5251, INRIA Project team Monc, Talence, France, F-33405 Talence Cedex, France}
\address{$^2$ INRIA Bordeaux Sud Ouest, 200 Av. de la Tour, Talence, 33405, France}

\eads{\mailto{eloise.inacio@inria.fr}, \mailto{laurent.facq@math.u-bordeaux.fr}, \mailto{luc.lafitte@inria.fr}, \mailto{clair.poignard@inria.fr}, \mailto{bdenisde@math.u-bordeaux.fr}}

%\author{E Inacio$^1$, L Lafitte, L Facq, C Poignard, B Denis de Senneville}

%\address{University of Bordeaux, IMB, UMR CNRS 5251, INRIA Project team Monc, Talence, France, F-33405 Talence Cedex, France}
%\address{INRIA Bordeaux Sud Ouest, 200 Av. de la Tour, Talence, 33405, France}
%\ead{eloise.inacio@inria.fr $^1$}

\vspace{10pt}
\begin{indented}
\item[]May 2024
\end{indented}

\begin{abstract}

\paragraph{Objective:} In medical imaging, it is often crucial to accurately assess and correct movement during image-guided therapy. Deformable image registration (DIR) consists in estimating the required spatial transformation to align a moving image with a fixed one. However, it is acknowledged that for DIR methods, boundary conditions applied to the solution are critical in preventing mis-registration. This poses an issue particularly when areas of interest are located near the image border. Despite the extensive research on registration techniques, relatively few have addressed the issue of boundary conditions in the context of medical DIR. Our aim is a step towards customizing boundary conditions to suit the diverse registration tasks at hand.
 
\paragraph{Approach:} We analyze the behavior of two typical global boundary conditions: homogeneous Dirichlet and homogeneous Neumann. We propose a generic, locally adaptive, Robin-type condition enabling to balance between Dirichlet and Neumann boundary conditions, depending on incoming/outgoing flow fields on the image boundaries. The proposed framework is entirely automatized through the determination of a reduced set of hyperparameters optimized via energy minimization.

\paragraph{Main results:} The proposed approach was tested on a mono-modal CT thorax registration task and an abdominal CT-to-MRI registration task. For the first task, we observed a relative improvement in terms of target registration error of up to 12\% (mean 4\%), compared to homogeneous Dirichlet and homogeneous Neumann. For the second task, the automatic framework provides results closed to the best achievable.

\paragraph{Significance:} This study underscores the importance of tailoring the registration problem at the image boundaries. In this research, we introduce a novel method to adapt the boundary conditions on a voxel-by-voxel basis, yielding optimized results in two distinct tasks: mono-modal CT thorax registration and abdominal CT-to-MRI registration. The proposed framework enables optimized boundary conditions in image registration without a priori assumptions regarding the images or the motion.
\end{abstract}

\vspace{2pc}
\noindent{\it Keywords}: Deformable Image Registration, Variational methods, Boundary conditions
%\maketitle

\maketitle

\section{Introduction}

%In medical imaging, multiple scans, often from different sensors, are required daily for a complete understanding of a patient's condition. For instance, an oncologist may want to estimate tumor growth before deciding on the best course of action, or an interventional radiologist may have to compare a previous scan with the current imaging to better guide a per-cutaneous ablation. However, because the images are not captured simultaneously, on the same patient, or, sometimes, within the same imaging modality, it is necessary to align features prior to comparing them. Indeed, motion artefacts (breathing, etc), changes in the patient (inter-patient registration, gaining/losing weight, etc) or position of the scanner relative to the patient(s) will make it laborious for the expert to correctly interpret and compare the images. Challenges also arise from the artifacts and noise introduced by the sensors themselves.

In medical imaging, numerous scans may be necessary on a daily basis to fully comprehend a patient's condition and/or plan treatment. This is particularly true for dose calculation, for instance in the case of radiotherapy \cite{RT_dose}, electroporation ablation \cite{elec_dose} or thermotherapy \cite{HIFU_dose}. As the images may not be captured simultaneously, or occasionally within the same modality, it becomes essential to align image features before drawing conclusions \cite{RTalign} \cite{DIRsurvey} \cite{HIFU_DIR}. Any motion event (such as breathing or peristalsis), or adjustment in the scanner's position relative to the patient, complicate the task for the expert in accurately interpreting and comparing the images.

%Registration is an image processing task consisting in estimating the motion to take one moving image onto one fix image by accounting for the content of both images and sometimes prior knowledge on the motion and/or the physical properties of the material. It has been an active area of research in the image processing field since the 1960s due to its wide variety of applications. To register images, various methods exist including parametric motion estimation \cite{param}, \cite{paramspline}, \cite{paramtmp}, variational methods \cite{EVolution}, \cite{demons}, \cite{cclg}, \cite{complex}, machine learning \cite{ML1}, \cite{ML2}, \cite{DL}, and combinations of any two \cite{mix}. They can be intensity driven or landmark driven. The former option has limitations in the case of multi-modality, even when extracted features such as SIFT \cite{sift} or SURF \cite{surf} are used: an organ of interest may be represented by very different intensities in a Magnetic Resonance Image (MRI) and a Cone Beam Computed Tomography (CBCT) scan for instance. Therefore, it becomes difficult to define a measure of similarity, necessary to guide the process and finally determine if the moving image was successfully registered onto the fix image.  The latter is difficult to use as the automatic generation of landmarks is still very challenging and manual landmarks are costly since they require expertise.

Deformable Image Registration (DIR) is an image processing task that consists in estimating the motion to align a moving image with a fixed image, taking into account the content of both images and sometimes prior knowledge of the motion and/or the tissue's physical properties. It has been an active area of research in the field of image processing since the 1960s, owing to its extensive range of applications. In pursuit of this goal, a variety of methods are available, including parametric motion estimation \cite{param} \cite{paramspline} \cite{paramtmp}, variational methods \cite{demons} \cite{cclg} \cite{complex} \cite{EVolution}, machine learning \cite{ML1} \cite{ML2} \cite{DL} or combinations \cite{mix}. DIR can be expressed as a variational method, offering a framework to formulate versatile DIR problems adaptable to various applications. The spatial transformation $\Tbf$ between a moving image $J$ and a fix image $I$ ($(I, J) \in \Omega \times \Omega$, $\Omega$ being the image domain: $\Omega \subset \mathbb{R}^{3} \rightarrow \mathbb{R}$ for the three dimensional case) is estimated by minimizing the following energy functional: 

\begin{equation}\label{eq:VarForm}
  \Ecal(\Tbf) = \int_{\Omega} (\Dcal^{I,J}(\Tbf) + \alpha \Scal(\Tbf)) \mathop{\textbf{dx}}
\end{equation}

\noindent where $\Tbf:\Omega \rightarrow \Omega$, $\Dcal$ is a data fidelity term which drives the motion so that relevant features are aligned between $I$ and the registered image $K=\Tbf(J)$, $\Scal$ is a regularization term which is crucial to achieve well-posedness \cite{rigid} \cite{generalReg} \cite{param} and $\alpha$ is a parameter allowing to balance between data fidelity and regularization. 

%In some cases, only a partial field of view is available. This may be inherent to the sensor’s nature or help reducing the acquisition duration if multiple scans need to be captured in a short amount of time. Here, an additional challenge arises as the voxels at the borders still convey relevant information but do not have the neighborhood necessary for an accurate estimation of the transformation. Similarly, when there is no sharp distinction between the features and the background, as in histological sections \cite{boundary}, background may be confused with relevant voxels and hinder the motion estimation. Often times, this issue is disregarded and, in the case of variational methods, simple boundary conditions are applied on every borders of the solution: namely homogeneous Dirichlet boundary conditions or homogeneous Neumann boundary conditions.

In certain scenarios, only a partial field of view is available, either due to inherent limitations of the imaging sensor or the acquisition process itself, when multiple scans need to be conducted within a limited timeframe. Consequently, an additional challenge arises for DIR methods, wherein the imposition of appropriate boundary conditions is crucial for preventing mis-registration from propagating towards the image center. This becomes particularly problematic when regions of interest lie near the image boundary. Several attempts restrict the registered area using segmentation masks either automatically extracted, provided by an extra module or, in some cases, by experts \cite{freeBC} \cite{FEBC}. Doing so allows shearing between the zone of interest and the background as well as limits the impact of the constraint applied on the boundary, since information is still available beyond. However, most of the time, the boundary conditions are implicitely imposed either with the homogeneous Dirichlet boundary condition (HD)--the motion is null at the border ($\textbf{T}(\textbf{x}) = 0$, $\forall \textbf{x} \in \delta \Omega$, $\delta \Omega$ being the image boundary)-- or the homogeneous Neumann boundary condition (HN) --the motion is constant along the normal border direction ($\nabla \textbf{T} \cdot \textbf{n} = 0$, $\textbf{n}(\textbf{x})$ being the vector normal to the image boundary at location $\textbf{x} \in \delta \Omega$)-- which is more adequate when objects are close to the edges \cite{boundary}.

An important feature, when solving the minimization problem for DIR, lies in the fact that the constraints on the motion on the boundary have to be {\it complementing} for the volume operator $\Scal$ so that the minimization problem is well-posed. With the standard regularization terms, this condition holds for Dirichlet and Neumann conditions, but it is not the case for any regularization, and the choice of the boundary conditions is tightly linked to the operator $\Scal$ of Eq. \eqref{eq:VarForm}.

Our contribution encompasses the following four key aspects:

\begin{enumerate}

%\item We first argue the importance of adapting the boundary conditions voxel-wise, especially in the case of partial fields of view where relevant information is also located at the edges of the image but lacks surrounding voxels for an adequate estimation of the motion. The choice of boundary condition is thus an hyper-parameter to the registration algorithm.

\item We first highlight the importance of tailoring boundary conditions, particularly in scenarios with partial fields of view where relevant information resides near the image border.% The selection of boundary conditions represents a hyper-parameter for the DIR algorithm.
    
%\item We propose a generic framework for voxel-wise adaptive boundary conditions in the form of a local boundary condition tending towards a homogeneous Dirichlet condition when there is no incoming/outgoing flow fields and relaxing the constraint otherwise.

\item We introduce a generic framework for voxel-wise adaptive boundary conditions in the form of a Robin-type local boundary condition tending towards a HD condition in the absence of incoming or outgoing flow fields, while relaxing the constraint under other circumstances.

%\item We introduce a method for the detection of flow fields on the edges, thus allowing to account for the specificity of the images in the boundary conditions.

\item We introduce a method for detecting and incorporating incoming or outgoing flow fields within the boundary conditions, thereby accommodating the image-specific characteristics.

\item We demonstrate that the proposed framework is capable of complete automation through the optimization of a reduced set of hyperparameters via DIR energy minimization.

\end{enumerate}

\section{Method}

\subsection{Generic DIR problem formulation}

In the present study, we focus on the standard Tikhonov regularization:

\begin{equation}
	\Scal(\Tbf)=\|\nabla \Tbf\|^2,
\end{equation} and the minimization of $\Ecal$ as expressed in Eq. (\ref{eq:VarForm}) is performed thanks to the Euler-Lagrange equations, for which the following Fourier-Robin condition (see Eq. \eqref{eq:Robin}) is known to be complementing.
More precisely, we focus on the following evolution problem on $\Tbf=[\Tbf_x,\Tbf_y,\Tbf_z]$ solved coordinate by coordinate, that is for $s\in\{x,y,z\}$, $\Tbf_s$ satisfies:
      
\begin{subequations}\label{eq:PBEv}
\begin{align}
    &\partial_t \Tbf_s -\alpha\Delta \Tbf_s =-\frac{\delta\Dcal^{I,J}}{\delta\Tbf_s}(\Tbf), \text{in $\Omega$}\hspace{2cm} \textrm{(Minimization of $\Ecal$)}\\
&\Tbf_s({t=0},x)=\Tbf^{init}_s(x),\quad x\in\Omega \hspace{2.5cm} \textrm{(Initial conditions)}\\
& \Abf_s(\nabla\Tbf_s\cdot \nbf) +\Bbf_s\Tbf_s =\Cbf_s, \text{ on $\partial\Omega$} \hspace{1.9cm} \textrm{(Boundary conditions)}\label{eq:Robin}
\end{align}
\end{subequations}

\noindent where $\Tbf^{init}$ is a prior vector field (in this study, $\Tbf^{init}$ was identically equal to zero), $\Abf,\Bbf, \Cbf$ are vectors --- $\Abf$ and $\Bbf$ having positive components $\Abf_s, \Bbf_s$ --- defined in the subsequent section. In Eq. \eqref{eq:Robin}, $\nbf$ refers to the normal vector to the boundary of the image, or more precisely, to the boundary of the field of view of interest. In fluid mechanics, Dirichlet boundary conditions are interpreted as an imposed velocity field on the image boundary, whereas Neumann boundary conditions imply the imposition of the shear tensor along the normal direction \cite{FM}. The Fourier-Robin boundary condition offers a way to balance between these two configurations.

\subsection{Proposed adaptive local boundary condition}

We proposed to drive the choice of the above vectors $\Abf,\Bbf$ and $\Cbf$ by the following considerations:

\begin{enumerate}

\item A balance between Dirichlet and Neumann conditions is achievable on a voxel-by-voxel basis.

\item The source term $\Cbf$ can enforce values for both incoming and outgoing flow fields in the image boundary.

\end{enumerate}

Let $\textbf{g}$ represent a voxelwise incoming/outgoing field map in the image boundary, we propose to choose $\Abf_s,\Bbf_s,\Cbf_s$ as follows:

\begin{equation}
\Abf_s=\beta (\textbf{g}_s),\qquad \Bbf_s=1-\Abf_s, \qquad \Cbf_s= \gamma \textbf{g}_s
\label{eq:ABC}
\end{equation}

\noindent where  $\gamma$ is a weighting factor for the source term and $\beta:\mathbb{R} \rightarrow [0, 1]$ is the 1-parameter sigmoidal function between 0 and 1 defined by:

\begin{equation}
    \forall \lambda \in \mathbb{R},\qquad \beta (\lambda) = \left\{ 
      \begin{array}{ c l }
        1 - \tanh(a-\vert \lambda \vert) & \quad \textrm{if } \vert \lambda \vert \le a\\
        1                 & \quad \textrm{otherwise}
      \end{array}
    \right.
\label{beta_func}
\end{equation}

\noindent $a$ being the second hyperparameter that tunes the transition between   Neumann ($\beta=1$) and Dirichet ($\beta=0$) conditions. 
Therefore, our proposed adaptive Robin condition (hereafter referred to as ``Adaptive Robin'') relies solely on the determination of two hyperparameters: $a$ and $\gamma$. 

%\indent For this boundary condition, the complete variational problem is:

%\begin{equation}
%    E(T_i) = \int_{\Omega} (D_i(I, \Tbf(J)) + \alpha S(T_i) \textbf{dx} + \alpha \int_{\delta \Omega} (\frac{1-\beta_i(\textbf{g})}{\beta_i(\textbf{g})}T_i - \gamma g_i)^2\textbf{dx}
%\end{equation}

\subsection{Estimation of incoming/outgoing flow fields on the image boundaries}

At this point, we need to compute an estimation of the incoming or outgoing flow fields at the image boundaries (\emph{i.e.} $\textbf{g}$), in order to implement the adaptive boundary conditions described in Eq. \eqref{eq:PBEv} and \eqref{eq:ABC}. We recall that this mapping enables the reduction of the comprehensive parameterization of the proposed voxelwise boundary conditions to the determination of two hyperparameters: $a$ and $\gamma$.

In this study, the proposed adaptive framework described in Eq. \eqref{eq:ABC} use an estimation of incoming/outgoing flow fields represented as a motion inverse consistency map computed on the image boundary. The motion inverse consistency relies on the hypothesis that DIR is a symmetric task, that is, switching the fix and the moving images will not impact the solution:

\begin{equation}
    \textbf{T}_{moving \rightarrow fix} = (\textbf{T}_{fix \rightarrow moving})^{-1}
\end{equation}

However, in practice, the images order may change the outcome due to asymmetry either in the problem formulation or in the numerical methods. The inverse consistency map $\textbf{g}$ is a vector-valued map which was computed as follows:

\begin{equation}
    \textbf{g}(\textbf{T}, \textbf{T}_{inv}) = \frac{\textbf{T} + \textbf{T}_{inv}(\textbf{T}) + \textbf{T}_{inv} + \textbf{T}(\textbf{T}_{inv})}{2}s, \text{ on $\partial\Omega$}
    \label{invC}
\end{equation}

\noindent where $\textbf{T} = \textbf{T}_{moving \rightarrow fix}$ and $\textbf{T}_{inv} = \textbf{T}_{fix \rightarrow moving}$. The calculation of $\textbf{g}$ with Eq. \eqref{invC} was restricted to voxels located on the image boundary $\delta \Omega$. 

The determination of $\textbf{g}$ involved the computation of forward and backward motion estimates (\emph{i.e.} $\textbf{T}$ and $\textbf{T}_{inv}$). This process being computationally expensive, $\textbf{g}$ is only estimated once per resolution and over a small number of iterations. Given the unknown nature of the correct boundary condition at this stage, a HN boundary condition (representing the least restrictive global boundary condition) was employed.

\subsection{Implemented DIR algorithm}

To assess the effectiveness of the suggested adaptive boundary condition, we solved the problem outlined in Eq. \eqref{eq:PBEv} employing the parameter selection specified in Eq. \eqref{eq:ABC}. The overall framework is summarised in pseudo-code in Appendix C. The solution was obtained using the following multi-modal data fidelity term introduced in \cite{EVolution}:

\begin{align}
    \label{data_fidelity}
   \Dcal^{I,J}(\Tbf)=\exp(-\Ccal(I, \Tbf(J))),\intertext{where}
   \Ccal(I, \Tbf(J)) = \frac{\int_\Gamma \vert \vec{\nabla}I\cdot\vec{\nabla}\Tbf(J)\vert \mathop{\textbf{dx}}}{\int_\Gamma \|\vec{\nabla}I\|_2\|\vec{\nabla}\Tbf(J)\|_2 \mathop{\textbf{dx}}},
   \end{align}
   
The data fidelity term $\mathcal{D}^{I,J}$ relies on the image gradient, enhancing the method's resilience to contrast reversals. Specifically, it facilitates alignment of both parallel and anti-parallel edges by considering only the absolute values of gradients. In the aforementioned equation, $\Gamma$ represents a three-dimensional patch centered around the voxel under consideration. Patch-based computation increases robustness against substantial structural alterations, encompassing transient objects among other variations. %The regularization adopts a diffusion-based approach.

This variational algorithm thus incorporates two hyper-parameters: (i) the patch size (small values leads to good performances but a patch size of one voxel induces instabilities in the numerical scheme ; in the following experiments, we use a patch-size of $5 \times 5 \times 5$), (ii) the regularisation factor $\alpha$ (it balances between the data fidelity term and the regularization in Eq.\eqref{eq:VarForm}. $\alpha$ below 0.1 leads to an unstable numerical scheme, above that value, the algorithm is fairly robust to changes in $\alpha$ ; we use a value of 0.4 going forward). A complete parameter study was done in \cite{EVolution} and we used those results to guide our choice of parameters.

\begin{figure}[h!]
\begin{minipage}[b]{\linewidth}
\centering 
\centerline{\includegraphics[width=\linewidth]{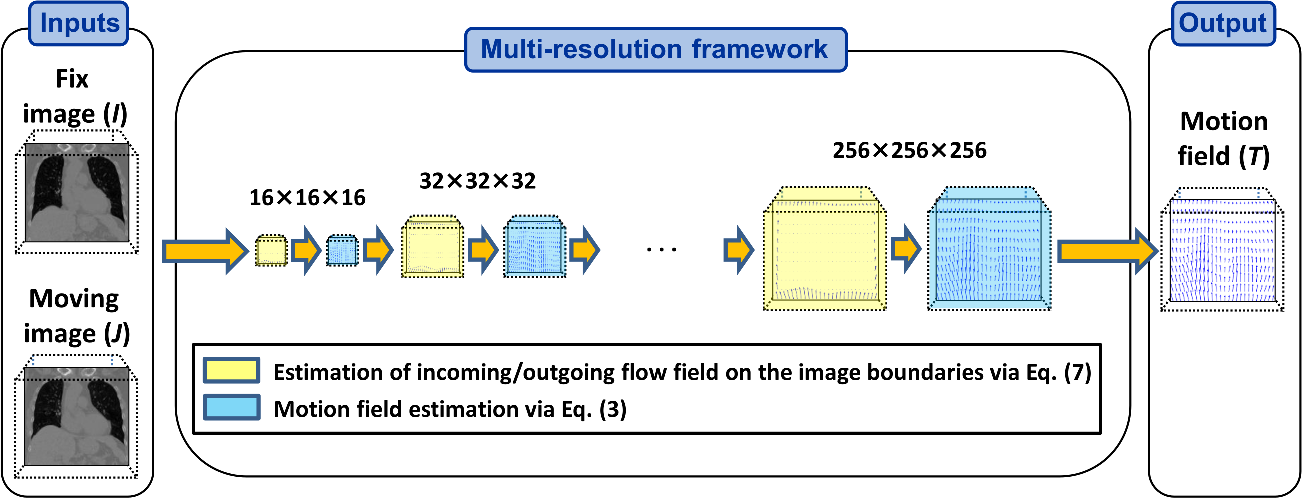}}
\end{minipage}
\caption{Generic multi-resolution framework for DIR incorporating proposed voxel-wise adaptive boundary conditions. An illustrative input image dimension of $256 \times 256 \times 256$ is displayed as a typical example. The associated pseudo-code is detailed in Appendix \ref{app:code}. The minimization of the energy $\mathcal{E}(\mathbf{T})$ was conducted on down-sampled versions of the images, progressively refining until reaching the original image dimension (blue blocks). The computation of incoming/outgoing flow fields on the image boundaries (\emph{i.e.} $\textbf{g}$) underwent updates throughout the registration process, at the beginning of each resolution step (yellow blocks). The system of equations (Eq. (\ref{eq:PBEv}a) and (\ref{eq:PBEv}b)) was solved in inner voxels (which corresponds to $\Omega \setminus \partial\Omega$), the one voxel wide perimeter (which corresponds to $\partial\Omega$) being computed separately according to the boundary conditions (Eq. (\ref{eq:PBEv}c), see Appendix \ref{app:BC_AR}).}
\label{fig:recap}%
\end{figure}

The general framework employed for adapting the boundaries voxel-wise during the registration process is shown in Figure \ref{fig:recap}. A multi-resolution methodology was employed, wherein energy minimization was conducted on down-sampled versions of the images, progressively refining until reaching the original image dimension. The computation of incoming/outgoing flow fields on the image boundaries (\emph{i.e.} $\textbf{g}$) underwent updates throughout the registration process, at the beginning of each resolution step, to ensure voxel-wise adaptation of boundary conditions. In the following experiments, the inplane dimension at lowest resolution level was $16 \times 16$ voxels (in-plane), the time step $dt$ was set to 1, the number of iterations for the calculation of $\textbf{g}$ via Eq. (\ref{invC}) was 10, the number of iterations for the minimization of $\mathcal{E}(\mathbf{T})$ via Eq. (\ref{eq:PBEv}) was 1000.

To properly apply the boundary conditions, the system of equations (Eq. (\ref{eq:PBEv}a) and (\ref{eq:PBEv}b) was solved in inner voxels (which corresponds to $\Omega \setminus \partial\Omega$), the one voxel wide perimeter (which corresponds to $\partial\Omega$) being computed separately according to the boundary conditions (Eq. (\ref{eq:PBEv}c), see Appendix \ref{app:BC}). 

\subsection{Hyperparameter optimization}

The two hyperparameters for the proposed adaptive Robin boundary conditions (\emph{i.e.}, $\gamma$ and $a$) were determined through a grid search process aimed at optimizing the minimized energy $\mathcal{E}(\mathbf{T})$ as expressed in Eq. \eqref{eq:VarForm}. The grid search spanned values of $\gamma \in [0, 1]$ with a step size of 0.1, and $a \in \{0, 0.1, 1, 10, 100\}$.

\subsection{Experimental setup}

\subsubsection{Mono-modal CT thorax registration:}

The first dataset employed for evaluating our methodology is the DIR-Lab 4DCT dataset \cite{Dir}. This dataset comprises Computed Tomography (CT) scans of lung tissue from patients undergoing treatment for esophageal cancer, thereby constituting a mono-modal registration task. From this database, we selected 10 pairs of CT scans: one acquired during inhalation and the other during exhalation. This dataset presents several challenges stemming from respiratory motion. Significantly larger displacements are evident near the diaphragm in comparison to other regions of the scan, alongside discontinuities at the interface between the lungs and the rib cage. To address this, we designate the scan acquired at extreme inhalation as the fixed image and the scan obtained at extreme exhalation as the moving image.

Each scan includes 300 manually placed landmarks within the lung regions, annotated by domain experts. These landmarks facilitate the computation of an error map across sparse voxels throughout the lungs, enabling localized assessment of the influence of boundary conditions on the estimation of the motion field.

%In the data set, scans have various slice resolution ranging from 0.97mm to 1.16mm as well as different magnitude of displacement: from 4.01mm on average in Case 1 up to 15.16mm on average in Case 8. We thus expect larger errors in the latter case.

\subsubsection{Multi-modal CT to MRI abdomen registration:}

The efficacy of the proposed adaptive boundary condition was also assessed in the first task of the Learn2Reg challenge in 2021. This task involved intra-patient CT to MRI multi-modal abdominal registration. The dataset comprised a total of 122 CT and MRI scans, with only 16 of them being paired. Each scan was associated with up to four segmentation masks delineating the liver, both kidneys, and the spleen. As not all paired scans had complete ground-truth segmentation available, seven pairs were selected for the experiment, ensuring inclusion of masks for all specified organs. The fixed and moving images corresponded to MRI and CT scans, respectively. Prior to registration, artificial paddings were eliminated via image cropping.

%The main challenges of this data set lies the multi-modal nature of the data and the ground-truth in the form of segmentation masks. They both make the evaluation more difficult as finding an adequate multi-modal metric is still an open question to compare the image and most geometric distances on segmentation masks miss key features.

\subsubsection{Assessment of the motion estimation process}

The registration outcomes derived from employing HD, HN, and the suggested adaptive Robin boundary condition (Eq. \eqref{eq:Robin}) are assessed as follows.

\paragraph{Mono-modal CT thorax registration:} 

%Registration was applied to the 10 CT scan pairs from the DIR-Lab data set.  \\
%\indent To further challenge the proposed framework, the same images are also cropped before registration. A margin of 10 voxels, corresponding to between 0.97 cm and 2.5 cm depending on the scan and the direction, was left around the landmarks furthest to the center.\\

%We tested two commonly used boundary conditions: homogeneous Dirichlet (equation \ref{Dirichlet}) and homogeneous Neumann (equation \ref{neumann}) ; and compare them to the proposed adaptive boundary condition: adaptive Robin (equation \ref{supRobin}) with vectorial (equation \ref{invC}) guidance. For evaluation, we compute 

First, the target registration error ($\TRE$) was computed using the provided 300 gold-standard landmarks in the following manner:

\begin{equation}
\label{eq:TRE}
    \TRE = \| \overrightarrow{r_I} + (\Tbf_x(\overrightarrow{r_I}), \Tbf_y(\overrightarrow{r_I}), \Tbf_z(\overrightarrow{r_I})) - \overrightarrow{r_J}\|_2
\end{equation}

\noindent where $\overrightarrow{r_I}$ is the coordinate of the landmark on the fix image, $\overrightarrow{r_J}$ is the coordinate of the landmark on the moving image and $\Tbf_x$, $\Tbf_y$, $\Tbf_z$ are the displacements in each direction, solution to Eq. \eqref{eq:PBEv}.

Following this, for each patient, the $\TRE$ computed for the 300 landmarks was interpolated using tri-linear interpolation onto a regular grid aligned with the image domain. Subsequently, a mean projection in the front-back direction of the resultant error map was computed to generate a 2D MIP error map.

%Lastly, to analyze the spatial disparities among motion fields acquired under the tested boundary conditions, the Euclidean distance was computed voxel by voxel between motion fields obtained using (i) HD and HN, (ii) HN and adaptive Robin, and (iii) HD and adaptive Robin boundary conditions.

\paragraph{Multi-modal CT to MRI abdomen registration:} 

In this experiment, we made use of the provided segmentation masks: we computed the average of the $Dice$ coefficients obtained for the four masks available:

\begin{equation}
    Dice(A, B) = \frac{2(A \cup B)}{\vert A \vert + \vert B \vert}
    \label{eq:dice}
\end{equation}

\noindent where $A$ represents the ground-truth segmentation mask on the fixed image, and $B$ represents the ground-truth segmentation mask on the moving image transformed with the estimated motion field. A $Dice$ value approaching 1 indicates a strong overlap between the masks.

%However, it is known that this kind of volume-based metrics often misses clinically relevant details, especially at the delineation of the masks \cite{DL}. Our method mainly impacting the edges of the image and thus the edges of the segmentation masks, we do not expect the impact of the adaptive boundary conditions to be as noticeable in the Dice coefficients on the second data set as in the TRE on the first data set.

\subsection{Implementation and Hardware}

Our experimental setup comprised a 24-core AMD Zen2 EPYC 7402 CPU paired with an Nvidia A100 GPU with 16 GB of memory. We coded our implementation utilizing Python along with the Cupy library \cite{cupy} for GPU optimization.

\section{Results}

\subsection{Mono-modal CT thorax registration}

As depicted in Table \ref{tab:DIRlab_TRE}, the proposed adaptive local boundary condition exhibited enhanced average $\TRE$ in contrast to HD and HN conditions ($\approx 4 \%$ relative $\TRE$ decrease). Particularly notable are the differences in case 7, wherein the adaptive boundary condition yields a 12 \% relative $\TRE$ enhancement. 

\begin{figure}
\begin{minipage}[b]{0.31\linewidth}
\centering
\centerline{\footnotesize{Homogeneous Dirichlet (HD)}}\medskip
\centerline{\includegraphics[trim={0cm 0cm 0cm 0cm},clip,height=3.9cm]{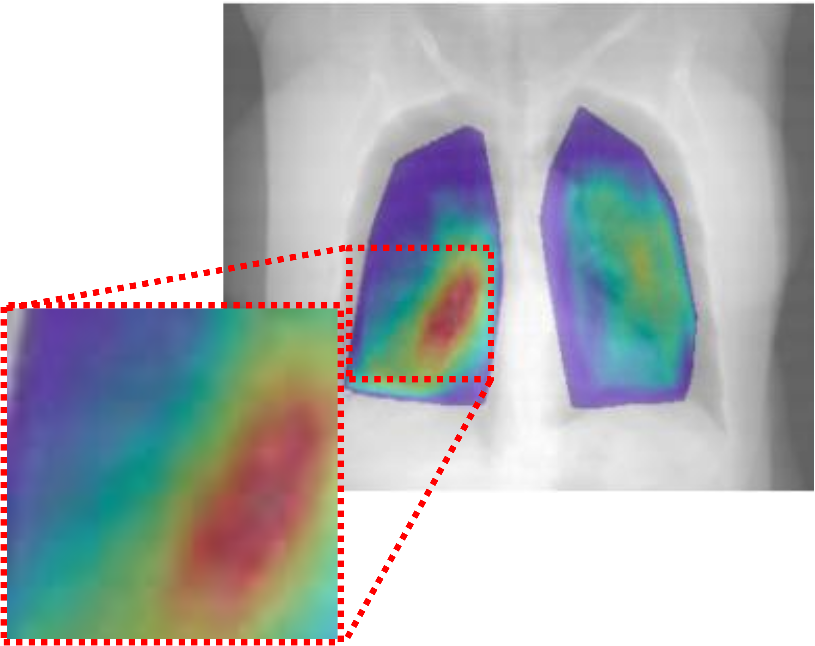}}
\centerline{(a)}\medskip
\end{minipage}
\begin{minipage}[b]{0.31\linewidth}
\centering
\centerline{\footnotesize{Homogeneous Neumann (HN)}}\medskip
\centerline{\includegraphics[trim={0cm 0cm 0cm 0cm},clip,height=3.9cm]{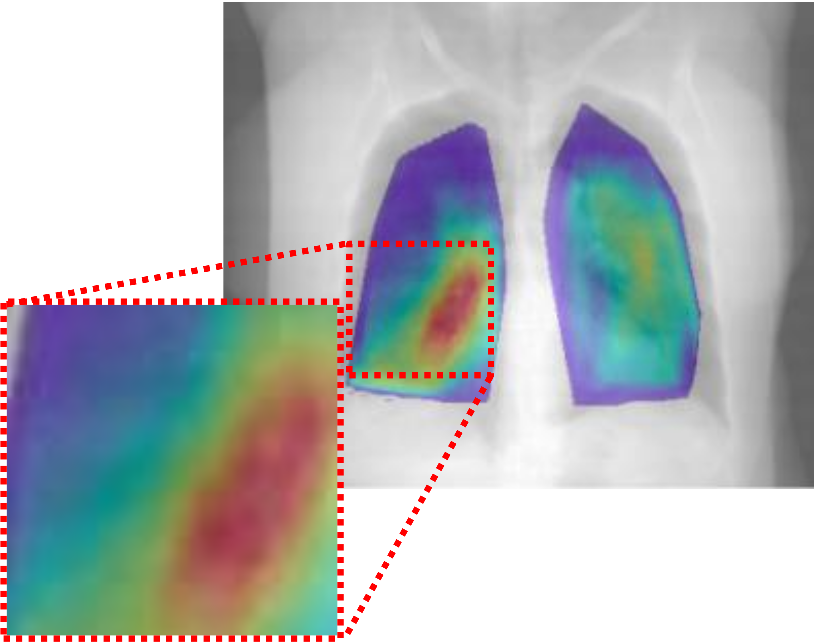}}
\centerline{(b)}\medskip
\end{minipage}
\begin{minipage}[b]{0.31\linewidth}
\centering
\centerline{\footnotesize{Adaptive Robin}}\medskip
\centerline{\includegraphics[trim={0cm 0cm 0cm 0cm},clip,height=3.9cm]{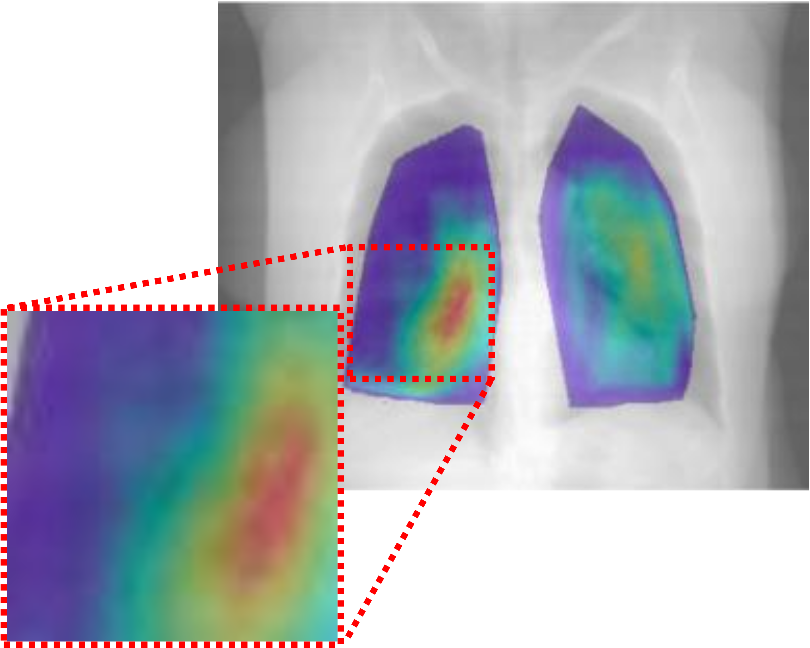}}
\centerline{(c)}\medskip
\end{minipage}
\begin{minipage}[b]{0.04\linewidth}
\centering
\centerline{}\medskip
\centerline{\includegraphics[trim={0cm 0cm 0cm 0cm},clip,height=3.9cm]{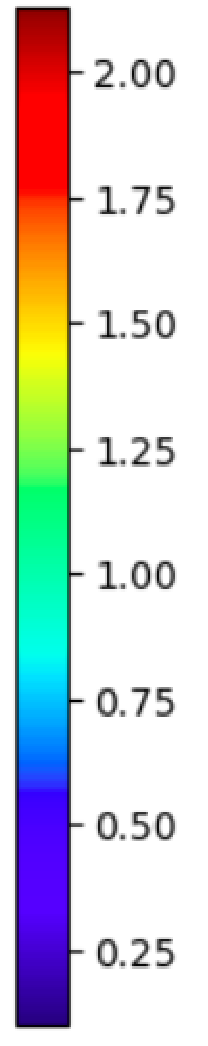}}
\centerline{}\medskip
\end{minipage}
\caption{Example of mono-modal CT thorax registration results (case 8). Mean intensity projection of error maps are reported using HN (a), HD (b) and the proposed adaptive Robin conditions (c) (colorbar in millimeters).}
	\label{mip}%
\end{figure}

Figure \ref{mip} provides mean projection of error maps obtained using HN, HD, and the proposed adaptive Robin conditions in case 8. The selection of boundary conditions resulted in variations of up to 8 mm. As anticipated, differences are primarily observable at the image boundary, particularly near the diaphragm where substantial upward motion occurs. Notably, some of these differences extend towards the image center, within the lungs, consequently influencing the $\TRE$ at landmarks (see Fig. \ref{mip}).

\begin{figure}
\begin{minipage}[b]{0.49\linewidth}
\centering
\centerline{\includegraphics[trim={0cm 0cm 0cm 0cm},clip,height=5.6cm]{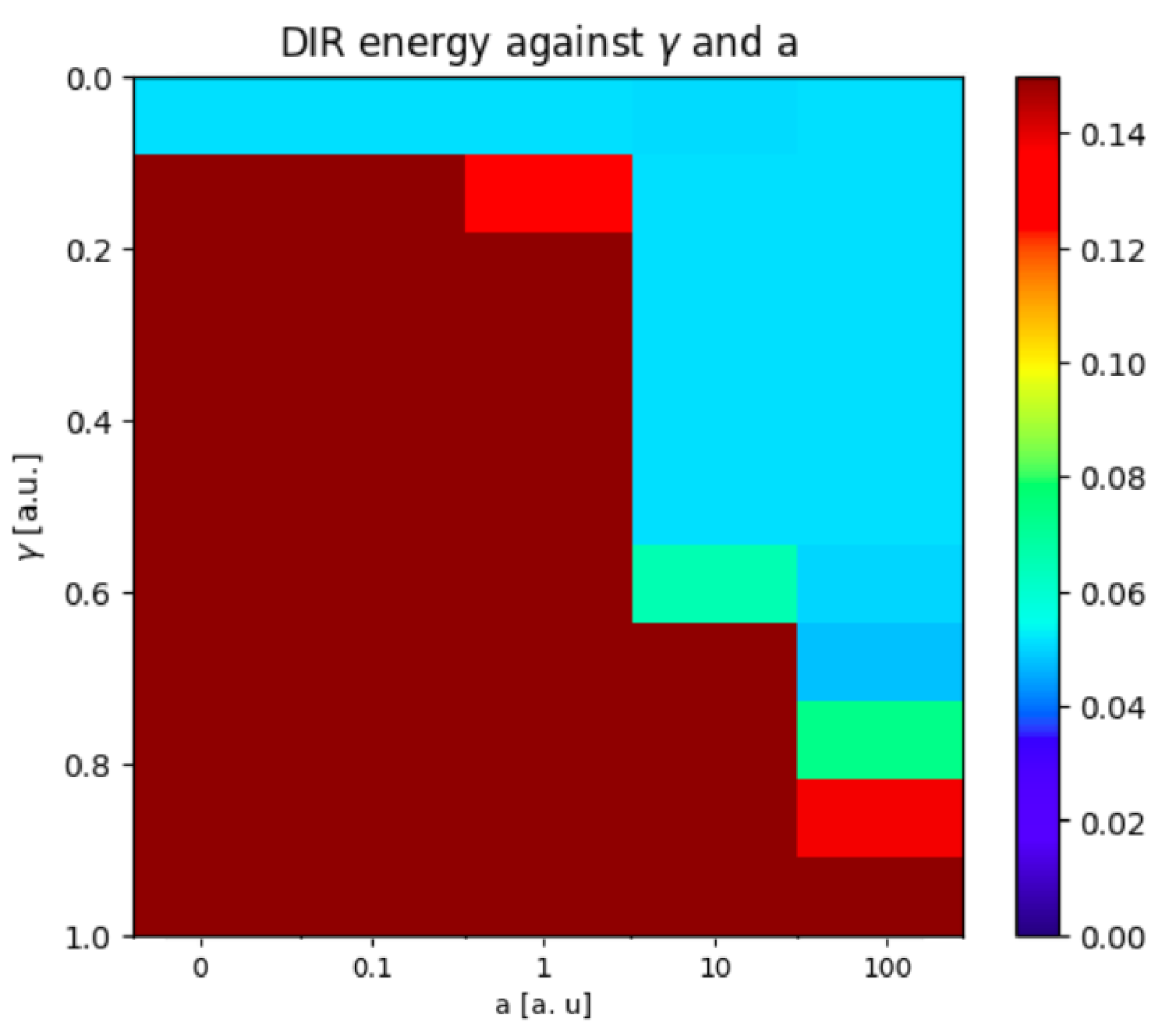}}
\centerline{(a)}\medskip
\end{minipage}
\begin{minipage}[b]{0.49\linewidth}
\centering
\centerline{\includegraphics[trim={0cm 0cm 0cm 0cm},clip,height=5.6cm]{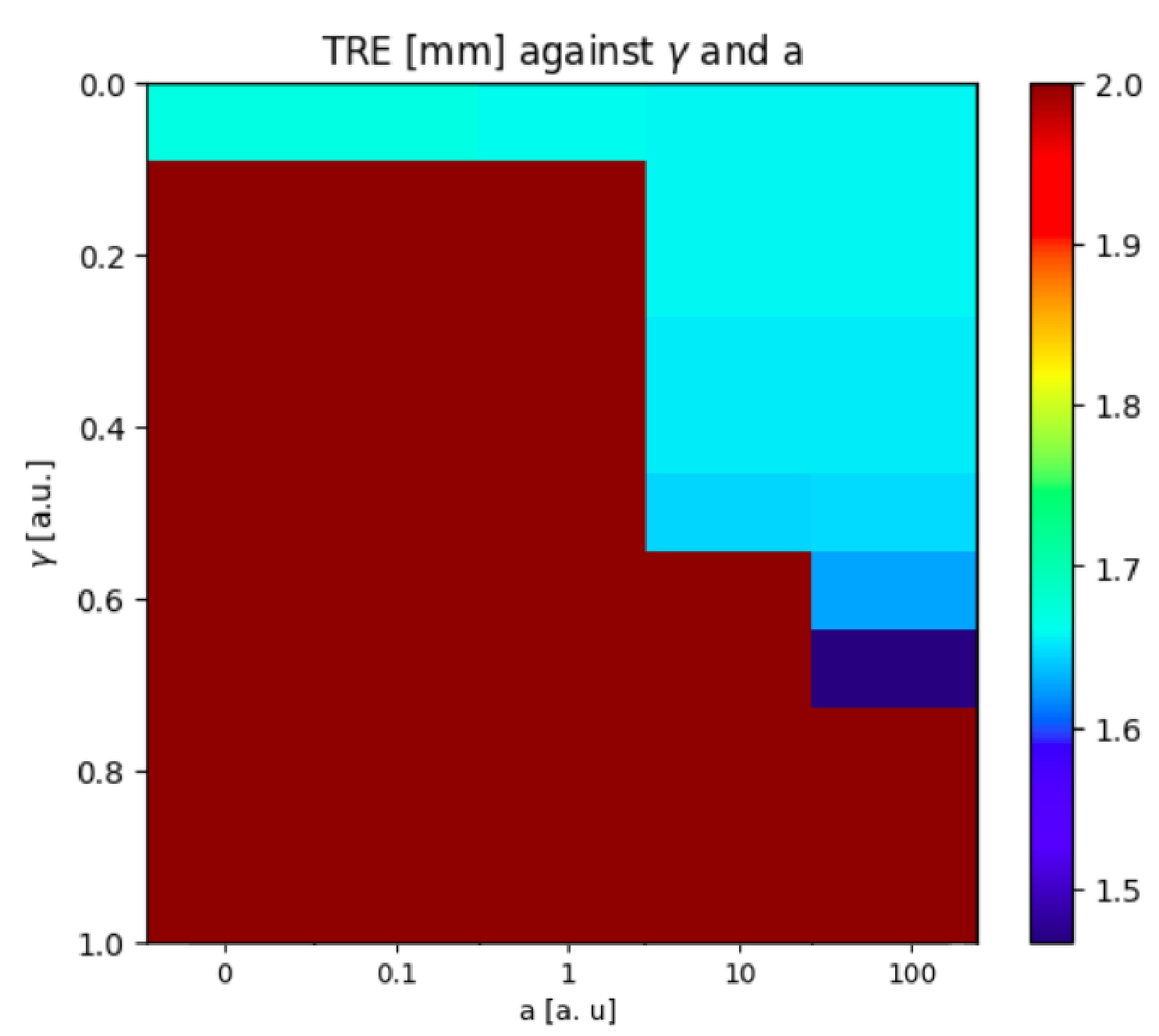}}
\centerline{(b)}\medskip
\end{minipage}
	\caption{Typical example of hyperparameter optimization (\emph{i.e.} $\gamma$ and $a$) for the proposed adaptive Robin boundary conditions (mono-modal CT thorax registration/case 7). (a): grid search optimizing the minimized DIR energy $\mathcal{E}(\mathbf{T})$ as expressed in Eq. \eqref{eq:VarForm} (minimum energy yields optimal hyperparameters); (b): $\TRE$ metric calculated using Eq. (\ref{eq:TRE}) for image registration assessment (minimum $\TRE$ yields optimal registration outcomes).} 
	\label{fig:grid_DIRlab}%
\end{figure}

A typical example of a hyperparameter optimization of the proposed adaptive local boundary condition is presented in Figure \ref{fig:grid_DIRlab}. The DIR energy $\mathcal{E}(\mathbf{T})$ (Fig. \ref{fig:grid_DIRlab}a), as defined in Eq. \eqref{eq:VarForm}, effectively aligns with the $\TRE$ metric used for image registration assessment (Fig. \ref{fig:grid_DIRlab}b). Please note that the DIR energy calculation was restricted to the 300 landmark locations for consistency with the computation of the Target Registration Error ($\TRE$) as defined in Equation \eqref{eq:TRE}. For cases 4 to 9, optimal identified hyperparameters consisted in a non-homogeneous Dirichlet boundary conditions ($a = 100$) using incoming/outgoing motion estimates as a source term ($\gamma$ around 0.6). This tailored parameterization underscores the necessity for substantially greater movement near the diaphragm compared to other image edges. Consequently, a non-homogeneous Dirichlet boundary condition enabling incoming/outgoing flow near the diaphragm was deemed appropriate.

Across the entire dataset, automatic hyperparameterization based on the DIR energy led to $\TRE$ that closely approached best achievable values (see extreme right two columns in Table \ref{tab:DIRlab_TRE}).

\begin{table}
    \caption{Mean $\TRE$ (mm) obtained for each case of the DIR-Lab data set (mono-modal CT thorax registration) and for each tested boundary conditions, with mean and standard deviation (SD). Best scores are highlighted with bold characters.}
    \label{tab:DIRlab_TRE}
    \centering
    \begin{tabular}{ccccc}
    \hline
    Case & Homogeneous & Homogeneous & \multicolumn{2}{c}{Adaptive Robin} \\ 
% & Neumann & Dirichlet &  & \\ 
 & Neumann & Dirichlet & \crule{2}\\
 & & & Automatized & Best achievable\\ 
    \hline
    1 & \textbf{1.05} & 1.06 & \textbf{1.05} & \textbf{1.05} \\
    2 & 1.02 & 1.02 & 1.02 & \textbf{1.01} \\ 
    3 & \textbf{1.20} & 1.23 & \textbf{1.20} & \textbf{1.20}\\ 
    4 & 1.43 & \textbf{1.42}  & \textbf{1.42} & \textbf{1.42}\\ 
    5 & 1.57 & 1.56 &  \textbf{1.51} & \textbf{1.51}\\ 
    6 & 1.49 & 1.53 & 1.42 & \textbf{1.41}\\ 
    7 & 1.67 & 1.68  & \textbf{1.47} & \textbf{1.47}\\ 
    8 & 4.56 & 4.61 & 4.39 & \textbf{3.83}\\ 
    9 & 1.26 & 1.26 & \textbf{1.24} & \textbf{1.24} \\ 
    10 & \textbf{1.71} & 1.73  & \textbf{1.71} & \textbf{1.71}\\ \hline
    Mean $\pm$ SD & 1.70 $\pm$ 1.04 & 1.71 $\pm$ 1.05 & 1.64 $\pm$ 0.99 & \textbf{1.59 $\pm$ 0.82}\\ 
    \hline
    \end{tabular}
\end{table}

%\begin{figure}
%	\centering 
%	\includegraphics[width=350pt]{dir7dist.eps}	
%	\caption{Typical Euclidean distance map (DIR-lab case 7) between motion fields obtained using homogeneous Dirichlet and homogeneous Neumann (a), local Robin and homogeneous Neumann (b), local Robin and homogeneous Dirichlet (c). Colorbar units in millimeters).}
%	\label{euclid}%
%\end{figure}

\subsection{Multi-modal CT to MRI abdomen registration:}

Table \ref{tab:l2r} presents a summary of $Dice$ scores achieved using the HD, HN, and the proposed adaptive local boundary condition across all patients in the dataset.

\begin{figure}
\begin{minipage}[b]{0.49\linewidth}
\centering
\centerline{\footnotesize{Fix image ($I$)}}\medskip
\centerline{\includegraphics[trim={0cm 0cm 0cm 0cm},clip,height=3.5cm]{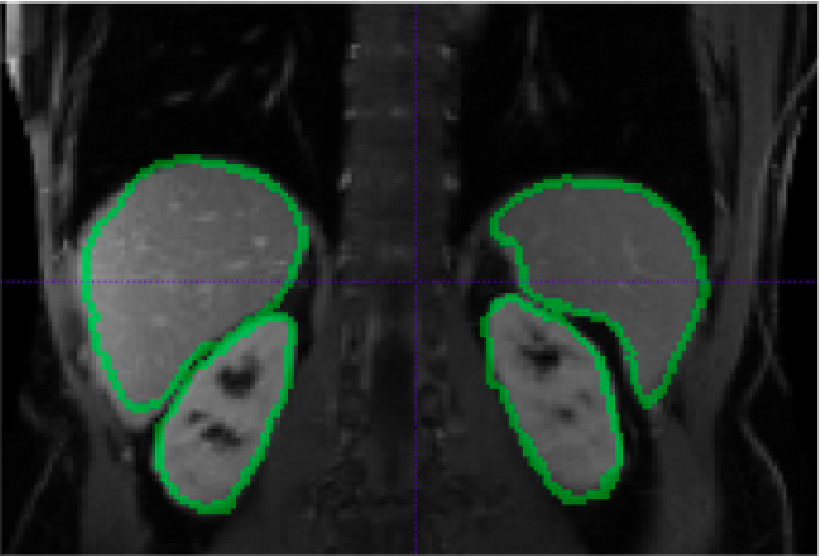}}
\centerline{(a)}\medskip
\end{minipage}
\begin{minipage}[b]{0.49\linewidth}
\centering
\centerline{\footnotesize{Moving image ($J$)}}\medskip
\centerline{\includegraphics[trim={0cm 0cm 0cm 0cm},clip,height=3.5cm]{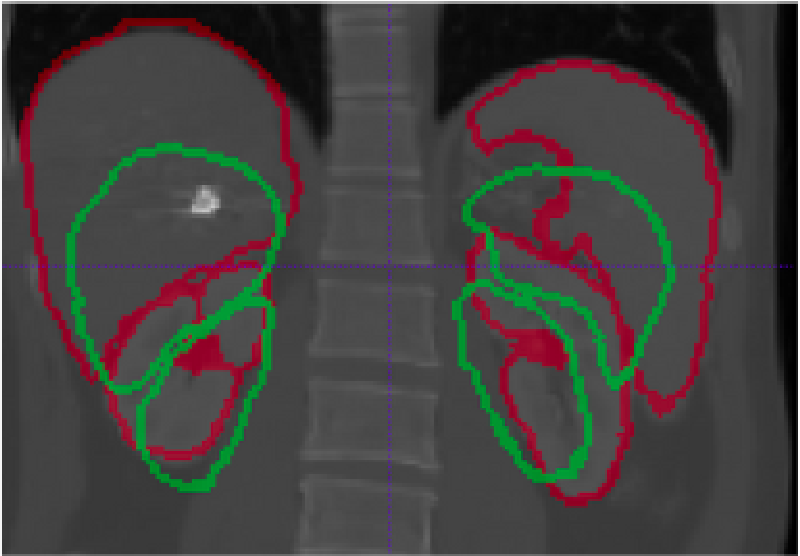}}
\centerline{(b)}\medskip
\end{minipage}
\begin{minipage}[b]{0.33\linewidth}
\centering
\centerline{\footnotesize{Homogeneous Dirichlet (HD)}}\medskip
\centerline{\includegraphics[trim={0cm 0cm 0cm 0cm},clip,height=3.5cm]{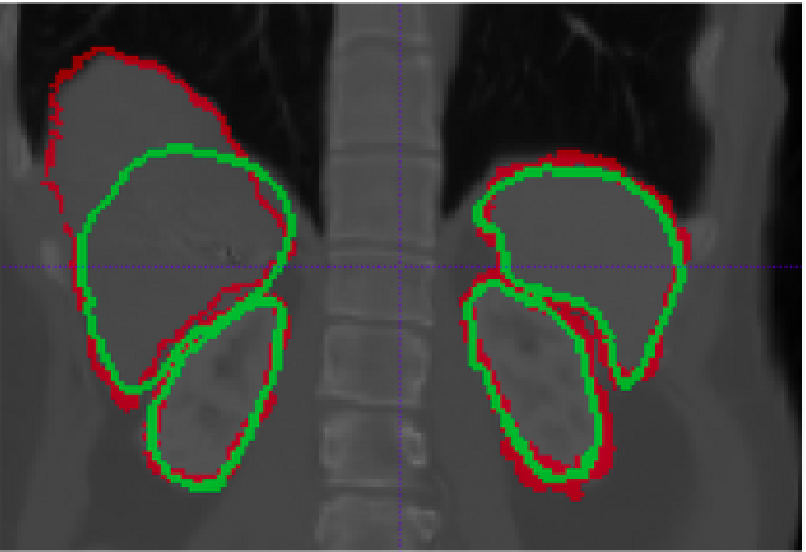}}
\centerline{(c)}\medskip
\end{minipage}
\begin{minipage}[b]{0.33\linewidth}
\centering
\centerline{\footnotesize{Homogeneous Neumann (HN)}}\medskip
\centerline{\includegraphics[trim={0cm 0cm 0cm 0cm},clip,height=3.5cm]{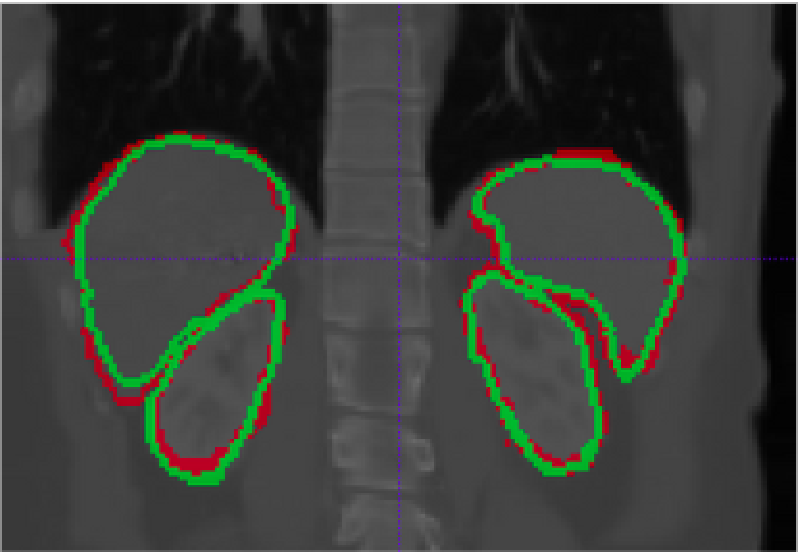}}
\centerline{(d)}\medskip
\end{minipage}
\begin{minipage}[b]{0.33\linewidth}
\centering
\centerline{\footnotesize{Adaptive Robin}}\medskip
\centerline{\includegraphics[trim={0cm 0cm 0cm 0cm},clip,height=3.5cm]{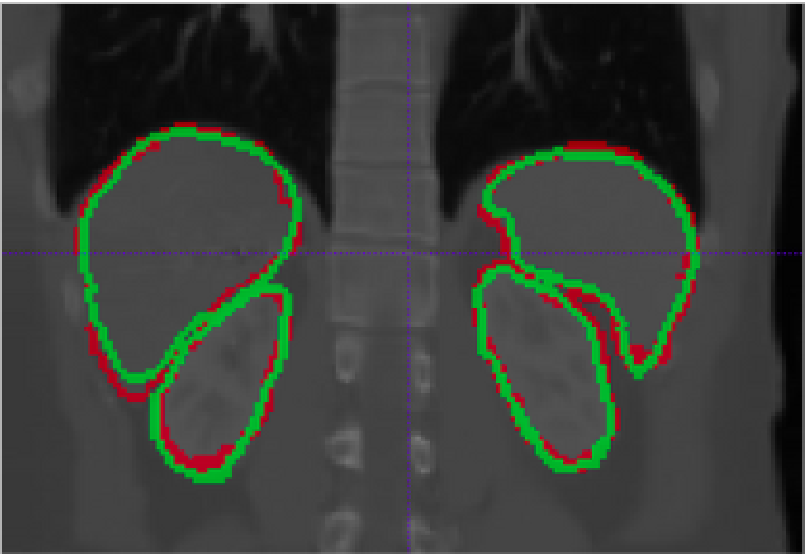}}
\centerline{(e)}\medskip
\end{minipage}
\caption{Example of multi-modal CT to MRI abdomen registration (case 4) showing the fix (a), moving (b) and registered images with HD (c), HN (d) and adaptive Robin (e). The segmentation contours are overlaid on all images to highlight the degree of alignment before and after registration (green: fix masks, red: registered masks). The labels include the liver, the spleen and both kidneys.}
	\label{fig:mask}%
\end{figure}

The HD boundary condition fails to effectively register the top section of the liver, as evidenced by the proximity of the red mask to its initial (unregistered) position (see Fig. \ref{fig:mask}c). These boundary inaccuracies propagate significantly towards the image center, particularly impacting the top left of the liver. This observation holds true for all patients within the dataset, consistently resulting in inferior outcomes with the Dirichlet boundary. Conversely, both the HN and adaptive Robin conditions yield similarly improved results.

\begin{figure}[!h]
\begin{minipage}[b]{0.49\linewidth}
\centering
\centerline{\includegraphics[trim={0cm 0cm 0cm 0cm},clip,height=5.6cm]{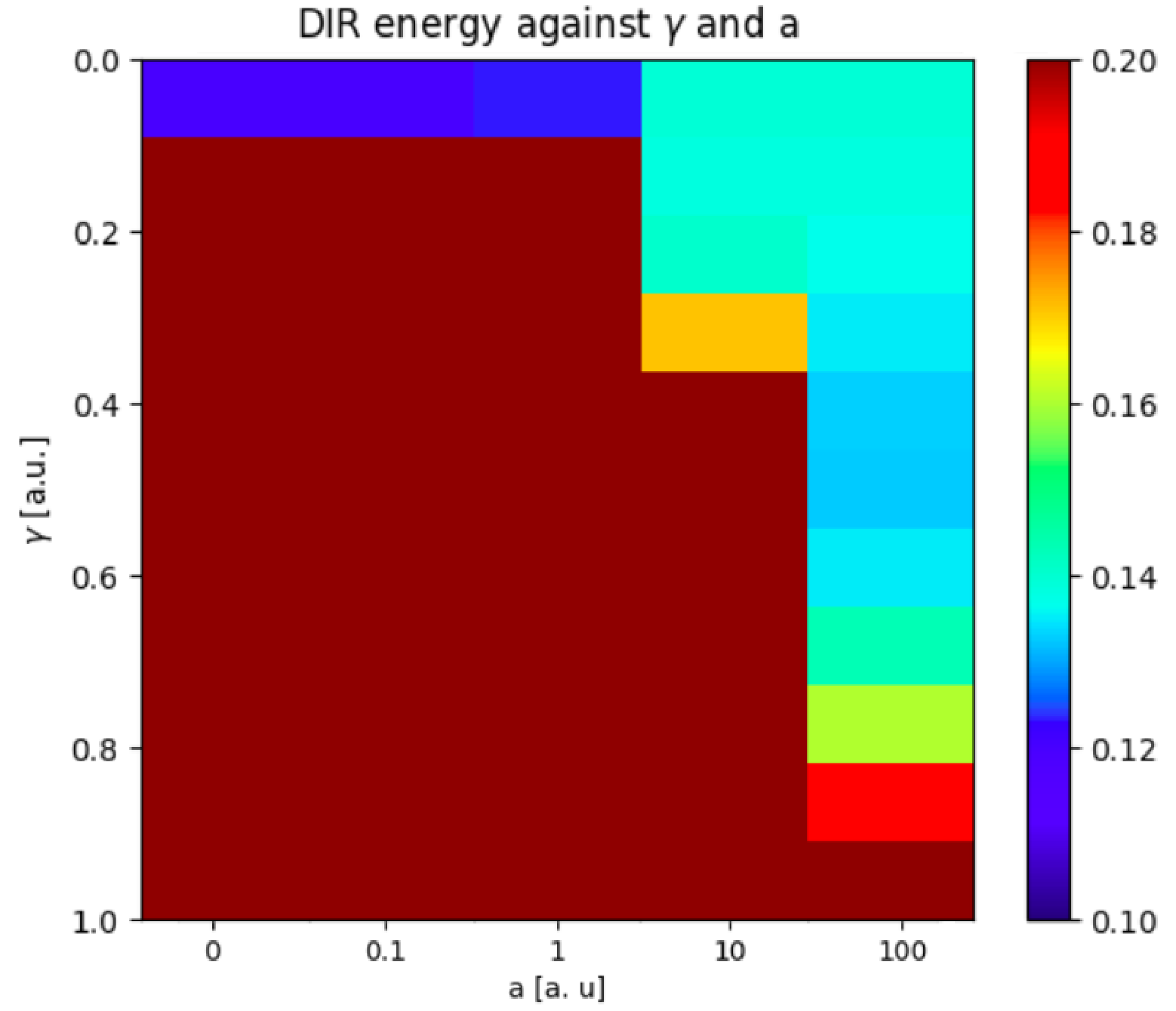}}
\centerline{(a)}\medskip
\end{minipage}
\begin{minipage}[b]{0.49\linewidth}
\centering
\centerline{\includegraphics[trim={0cm 0cm 0cm 0cm},clip,height=5.6cm]{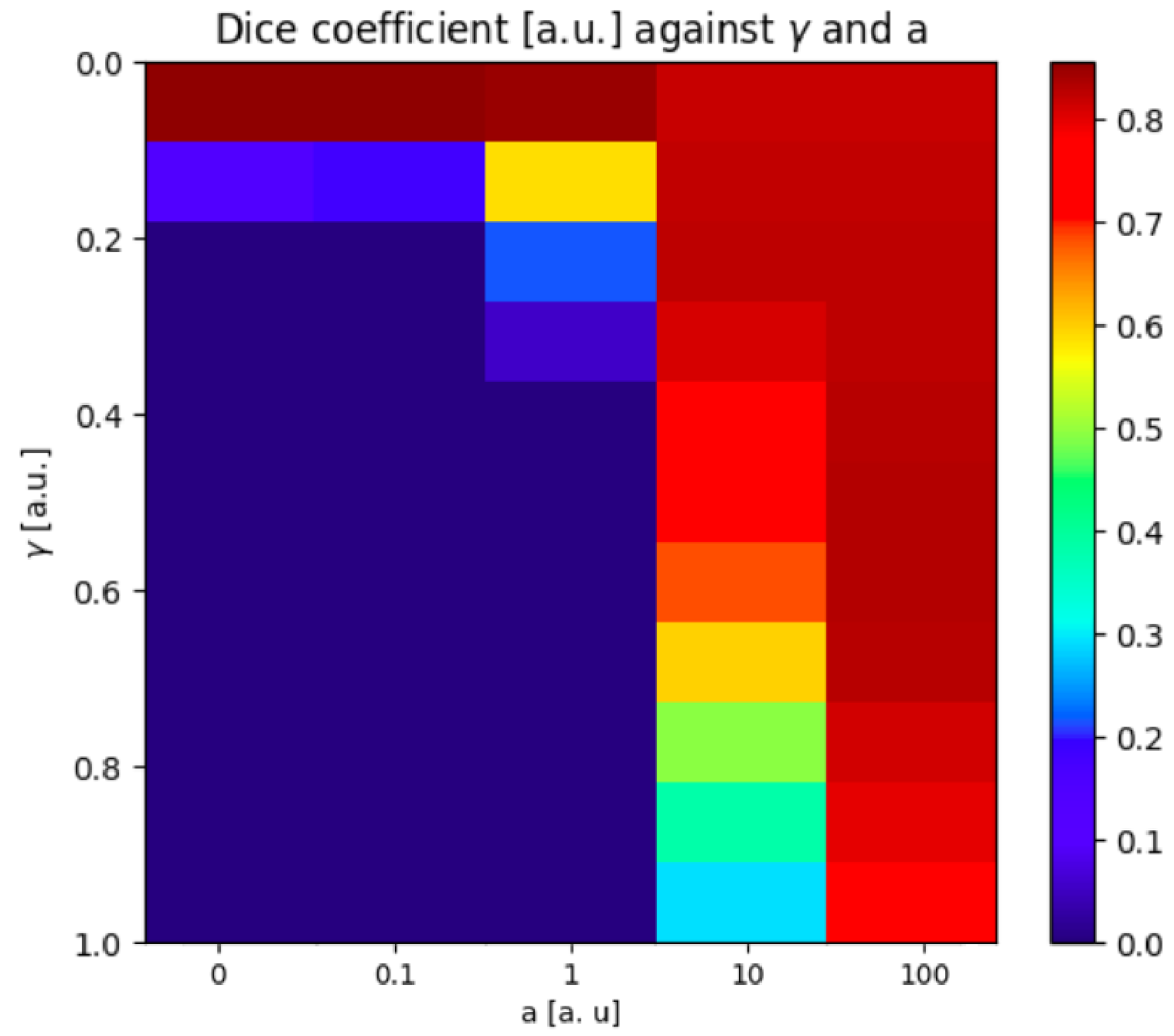}}
\centerline{(b)}\medskip
\end{minipage}
\caption{Typical example of hyperparameter optimization (\emph{i.e.} $\gamma$ and $a$) for the proposed adaptive Robin boundary conditions (multi-modal CT to MRI abdomen registration/case 4). (a): grid search optimizing the minimized energy $\mathcal{E}(\mathbf{T})$ as expressed in Eq. \eqref{eq:VarForm} (minimum energy yields optimal hyperparameters); (b): $Dice$ metric calculated using Eq. (\ref{eq:dice}) for image registration assessment (maximum $Dice$ yields optimal registration outcomes).}
	\label{fig:gridL2R}
\end{figure}

A typical example of hyperparameter optimization for the proposed adaptive local boundary condition is illustrated in Fig. \ref{fig:gridL2R} (case 4). The DIR energy (Fig. \ref{fig:gridL2R}a), as defined in Eq. \eqref{eq:VarForm}, closely corresponds to the gold standard $Dice$ values (Fig. \ref{fig:gridL2R}b). While the non-homogeneous Dirichlet boundary condition yielded satisfactory results with an average $Dice$ score around 0.80 (against 0.87 for best achievable), the optimized parameterization identified during grid search converges the adaptive local boundary condition towards a HN boundary condition ($\gamma = 0$ and $a = 0$). This customized parameterization emphasizes the necessity for significantly greater movement across all image edges.

\begin{table}
    \caption{Dice coefficients (averaged over 4 labels) obtained for each case of the Learn2Reg 2021 data set (multi-modal CT to MRI abdomen registration) and for each tested boundary conditions, with mean and standard deviation (SD). Best scores are highlighted with bold characters.} 
    \label{tab:l2r}
    \centering
    \begin{tabular}{ccccc}
    \hline
    Case & Homogeneous & Homogeneous & \multicolumn{2}{c}{Adaptive Robin} \\ 
% & Neumann & Dirichlet &  & \\ 
 & Neumann & Dirichlet & \crule{2}\\
 & & & Automatized & Best achievable\\ 
    \hline
        1 & \textbf{0.846} & 0.793 & \textbf{0.846} & \textbf{0.846}\\ 
        2 & \textbf{0.695} & 0.608 & \textbf{0.695} & \textbf{0.695}\\
        3 & 0.855 & 0.818 & \textbf{0.856} & \textbf{0.856} \\
        4 & \textbf{0.866}  & 0.760  & 0.859 & \textbf{0.866} \\ 
        5 & 0.722 & 0.556 & 0.722 & \textbf{0.756}\\ 
        6 & \textbf{0.810} & 0.529 & \textbf{0.810} & \textbf{0.810} \\ 
        7 & 0.846 & 0.829 & \textbf{0.849} & \textbf{0.849} \\ \hline
        Mean $\pm$ std &0.806 $\pm$ 0.069 & 0.699 $\pm$ 0.130 & 0.805 $\pm$ 0.068 & \textbf{0.811 $\pm$ 0.064}\\ \hline
    \end{tabular}
\end{table}

\section{Discussion}

To properly formulate a DIR problem applied to images acquired at distinct instants, two strategies are typically employed for handling motion in the image boundaries: either the velocity (Dirichlet) or the shear tensor along the normal (Neumann) are nullified. These global boundary conditions offer the advantage of being implementable using the discrete sine transform and the discrete cosine transform, respectively, leading to a significant reduction in computation time for CPU implementations.  Our research underscores the importance of tailoring the DIR algorithm specifically at image boundaries. A generalized approach guided by a reduced set of hyperparameters is proposed to address this concern. Additionnally, a straightforward and flexible numerical framework is introduced for adjusting boundary conditions during the registration process.

%The framework is fully automatised through energy minimisation. Both hyperparameters, namely $a$ and $\gamma$, are determined via grid search. When ground truth is available, the energy is computed on areas of interest: on the landmarks or within the segmentation masks. Experiments showed that, when computed on the full image, the parametrisation still provides results similar or close to those obtained on the ground truth exclusively. We thus argue that our approach would provide quasi-optimal registration when no prior information is available. Moreover, when minimum error and minimum energy are not given by the same parameters $(\gamma, a)$, the image registered with hyperparameters corresponding to the minimum energy is often more realistic that the image registered with the hyperparameters corresponding to the minimum errors, regardless of the dataset.

Presented results indicate that the suggested framework is capable of complete automation via DIR energy minimization. In both the mono-modal CT thorax registration and multi-modal CT to MRI abdomen registration tasks, the proposed adaptive boundary condition consistently produces superior or equivalent results when compared to the commonly employed global boundary conditions, namely HD or HN. The proposed framework customized the boundary conditions based on case-specific motion characteristics: on the one hand, within the mono-modal CT thorax dataset, significant motion is primarily concentrated near the diaphragm, with minimal motion elsewhere, hence a non-homogeneous Dirichlet boundary condition ; on the other hand, in the multi-modal CT to MRI abdomen dataset, minor movements are distributed throughout the image, reflecting its reduced coverage of the abdominal region, hence a HN boundary condition. The adaptive boundary condition still enables the transition to global boundary conditions when necessary, as evidenced in the multi-modal CT to MRI abdomen registration experiments. 

It was observed that the HD, HN, and non-homogeneous Dirichlet boundary conditions yielded the best results in the experiments. Conversely, the non-homogeneous Neumann conditions never produced optimal results. One possibility is that the source term $\Cbf$ is not well-suited for non-homogeneous Neumann conditions. However, since both HN and non-homogeneous Dirichlet conditions produced the best results, it appears that our choice of $\Cbf$ is generally satisfactory. Another potential explanation for the suboptimal performance of non-homogeneous Neumann boundary conditions could be the absence of applied shear stress within the field of view between the fixed and moving images in both tested datasets. In mini-invasive interventional procedures, such as radiofrequency \cite{RFA} or irreversible electroporation procedures \cite{elec_dose}, needles are introduced, generating local displacement due to shear stress in the normal direction. It can be anticipated that non-homogeneous Neumann boundary conditions may be more appropriate in such scenarios. Future studies on percutaneous ablations will further confirm or refute this assertion.

%These remarks also suggest that the observed hyperparameter space based for specific clinical application scenario.

In this paper, we estimated incoming and outgoing flow fields utilizing the motion inverse consistency metric. Our underlying hypothesis is that discrepancies between the forward and backward motion estimates are interpreted as material flow exiting or entering the field of view. The utilization of motion inverse consistency was driven by its practical advantage of relying solely on motion estimates. This characteristic makes this metric compatible with both mono- and multi-modal registration tasks, as it does not rely on image intensity. Furthermore, it does not require any additional information such as landmarks or image segmentations. Nevertheless, this method may result in erroneous estimation of incoming/outgoing flow fields in several scenarios. For instance, when there is limited or no tissue contrast present at the image boundary, false negative flows may be identified, as the registration process is predominantly influenced by the regularization term. Moreover, false positive flows may be detected in the presence of transient objects due to fluctuations in the data fidelity term. For these reasons, the motion inverse consistency metric may not be suitable in image regions involving needle insertions, as encountered in percutaneous minimally invasive interventional procedures. Application specific solution may be investigated thanks to the highly modular aspect of the proposed framework expressed in Eq. \eqref{eq:ABC}.

%{\color{red}Mettre temps de calcul pour une image du learn2reg (sans recherche des hyperparamètres)?}

%Those datasets only enable to highlight the benefits of homogeneous boundary conditions and non-homogeneous Dirichlet boundary conditions. It begs the following questions: are non-homogeneous Neumann boundary conditions not suitable for the kind of motion tackled in the DIR-Lab 4DCT dataset and the Learn2Reg dataset? or is our flow fields map $\textbf{g}$ not suited to a non-homogeneous Neumann boundary condition? These are to be investigated in future works. 

\section{Conclusion}

A DIR toolbox is essential in the field of medical imaging due to the growing need to align images of the same scene acquired across time and/or modalities. This study emphasizes the significance of customizing the DIR algorithm at image boundaries. A generalized formulation for this issue is presented, along with a direct and adaptable framework for tailoring boundary conditions throughout the registration process. Its adaptability stems from its capacity to shift between homogeneous Neumann, homogeneous Dirichlet, homogeneous Robin, and their non-homogeneous counterparts, conditioned by a reduced number of hyper-parameters. Additionally, we demonstrated the feasibility of fully automatizing such boundary conditions.  

The suggested local adaptive Robin boundary condition yields results closely approaching the best achievable outcomes for both examined mono-modal CT thorax registration and multi-modal CT to MRI abdomen registration tasks.

The impact of boundary conditions on more complex motion regularization term and different DIR formulations will need to be addressed in future works. Moreover, future studies will challenge the proposed method for other non-rigid multi-sensor registration scenarios (\emph{i.e.} CT to CBCT, MRI to CBCT, MRI to echography, etc). A crucial and forthcoming perspective will involve investigating the impact of dose delivery, particularly in procedures such as radiotherapy \cite{RT_dose}, thermotherapy \cite{RFA} and irreversible electroporation \cite{elec_dose}.

\section*{Acknowledgements}

This study has been partly granted by the Plan Cancer Projet MECI, n°21CM119-00. For the purpose of Open Access, a CC-BY public copyright licence has been applied by the authors to the present document and will be applied to all subsequent versions up to the Author Accepted Manuscript arising from this submission.

Experiments presented in this paper were carried out using the PlaFRIM experimental testbed, supported by Inria, CNRS (LABRI and IMB), Universit\'e de Bordeaux, Bordeaux INP and Conseil R\'egional d’Aquitaine (see https://www.plafrim.fr/).

\appendix

\section{Numerical implementation of the boundary conditions}
\label{app:BC}

\subsection{Homogeneous Dirichlet}
\label{app:BC_HD}

For homogeneous Dirichlet, we simply set the value 0 in the one voxel wide image perimeter (which corresponds to $\partial\Omega$).

%\subsection{Local Dirichlet boundary condition}

%The local Dirichlet boundary condition takes the form: 

%\begin{equation}
%    \Tbf = \gamma \textbf{g}
%\end{equation}

%\indent In this case, a straightforward multiplication of $\textbf{g}$ by the scalar $\gamma$ gives the right value for the motion field on the boundary. 

\subsection{Homogeneous Neumann}
\label{app:BC_HN}

\newcommand{\ds}{\mathrm{ds}}

To apply a homogeneous Neumann condition, $\nabla \textbf{T}_s \cdot \textbf{n}$ for $s\in\{x,y,z\}$ was computed using the following second-order approximation with a spatial step size $\ds$:

\begin{equation}
    \nabla \textbf{T}_s^0 \cdot \textbf{n} = - \frac{-3 \textbf{T}_s^0 + 4 \textbf{T}_s^1 - \textbf{T}_s^2}{2\ds}
\end{equation}

On the other side of the boundary, one have:

\begin{equation}
    \nabla \textbf{T}_s^N \cdot \textbf{n} = \frac{\textbf{T}_s^{N-2} - 4 \textbf{T}_s^{N-1} + 3 \textbf{T}_s^N}{2\ds}
\end{equation}

\noindent where the superscript gives the number of voxels along the direction considered.

Substituting into $\nabla \textbf{T}_s^{0, N} \cdot \textbf{n} = 0$ and solving for $\textbf{T}_s^0$ and $\textbf{T}_s^N$ respectively, we obtain: 

\begin{equation}
    \textbf{T}_s^0 = \frac{4 \textbf{T}_s^1 - \textbf{T}_s^2}{3}
\end{equation}

\begin{equation}
    \textbf{T}_s^N = \frac{4 \textbf{T}_s^{N-1} - \textbf{T}_s^{N-2}}{3}
\end{equation}

These conditions were imposed in the one voxel wide image perimeter (which corresponds to $\partial\Omega$).
%These must be verified on every edge, including at the image corners.

\subsection{Adaptive Robin boundary condition}
\label{app:BC_AR}

The adaptive Robin boundary condition takes the form: 

\begin{equation}
	\beta(\textbf{g}_s)(\nabla\Tbf_s\cdot\nbf)+(1-\beta(\textbf{g}_s))\Tbf_s = \gamma\textbf{g}_s
	\label{eq:robin}
\end{equation}

\noindent where $\textbf{n}$ is the normal to the boundary $\delta\Omega$, $\beta : \mathbb{R} \rightarrow [0, 1]$ is computed using Eq. (\ref{beta_func}), and $\gamma$ is one of the two hyper-parameters for the adaptive Robin boundary condition.

Making the same approximation as in Appendix \ref{app:BC_HN}, substituting in Eq. \eqref{eq:robin} and solving for $\textbf{T}_s^0$ and for $\textbf{T}_s^N$ with $\ds=1$:

\begin{equation}
    \textbf{T}_s^0 = \frac{\beta_s^0(4 \textbf{T}_s^1 - \textbf{T}_s^2) + 2\gamma \textbf{g}_s^0}{\beta_s^0 + 2}
\end{equation}

\begin{equation}
    \textbf{T}_s^N = \frac{\beta_s^N (4 \textbf{T}_s^{N-1} - \textbf{T}_s^{N-2}) + 2\gamma \textbf{g}_s^N}{\beta_s^N + 2} 
\end{equation}

These conditions were imposed in the one voxel wide image perimeter (which corresponds to $\partial\Omega$).
%These must be verified on every edge of the image, including the corners.

\section{Pseudo-code}
\label{app:code}

Algorithm \ref{alg:main_algo} corresponds to the primary framework for registration for typical $256\times256\times256$ images, encompassing the calculation of the voxelwise incoming/outgoing boundary flow designed to guide boundary conditions, the initialization of the motion field for the forthcoming estimation process at the current resolution, the registration algorithm, and the multi-resolution scheme (see Fig. \ref{fig:recap}).

\begin{algorithm}
\caption{Primary framework for registration}
         \SetKwInOut{Input}{Input}
        \SetKwInOut{Output}{Output}
\Input{$(I, J, \Tbf^{init})$ = (fix image, moving image, initial condition)}
\Output{$\Tbf$ = motion field}
$\Tbf \gets$ Resample $\Tbf^{init}$ to $(16 \times 16 \times 16)$\\
\For{$res \in \{(16 \times 16 \times 16), (32 \times 32 \times 32), ..., (256 \times 256 \times 256)\}$}{
$I' \gets$ Resample $I$ to $res$\\
$J' \gets$ Resample $J$ to $res$\\
$\Tbf \gets$ Resample $\Tbf$ to $res$\\
$\textbf{g}, \Tbf \gets$ Algorithm 2($I', J', \Tbf$)\\
%$\textbf{g} \gets$ Algorithm 2($I', J', \Tbf$)\\
$\Tbf \gets$ Algorithm 3($I', J', \Tbf, \textbf{g}$)\\
}
\label{alg:main_algo}
\end{algorithm}

Algorithm \ref{alg:InOutFlow} computes the incoming/outgoing flow field $\textbf{g}$ via Eq. (\ref{invC}). %, injected in the adaptive boundary condition later on from an initial motion estimated with non-homogeneous Neumann boundary conditions, as well as the initial motion field for the current resolution. 
It corresponds to the yellow blocks delineated in Figure \ref{fig:recap}. Note that $i$ is a local variable determing the iteration number (max=10).

\begin{algorithm}
\caption{Calculation of incoming/outgoing flow field in the boundary}
\SetKwInOut{Input}{Input}
\SetKwInOut{Output}{Output}
\Input{$(I, J, \Tbf)$ = (fix image, moving image, motion field)}
\Output{$\textbf{g}$ = Voxelwise in/out flow field in the boundary\\
$\Tbf$ = motion field}
$D \gets$ data fidelity term calculated from $I$, $J$ and $\Tbf$ using Eq. (\ref{data_fidelity})\\
$\Tbf_{inv} \gets \Tbf$\\
$J_{cur} \gets J$ transformed with $\Tbf$\\
%{\color{red}Est-ce que le terme d'attache aux données ne doit pas être updaté dans les itérations comme dans l'Algo 3?}\\
$i \gets 0$\\
\While{$i<10$}{
Update $\Tbf$ boundaries according to homogeneous Neumann boundary conditions (Appendix \ref{app:BC_HN})\\
$\Tbf \gets \mathcal{L}(I, J_{cur}, D, \Tbf)$ \\
$i \gets i + 1$\\}
$i \gets 0$\\
\While{$i<10$}{
Update $\Tbf_{inv}$ boundaries according to homogeneous Neumann boundary conditions (Appendix \ref{app:BC_HN})\\
$\Tbf_{inv} \gets \mathcal{L}(J_{cur}, I, D, \Tbf_{inv})$ \\
$i \gets i + 1$\\}
$\textbf{g} \gets$ incoming/outgoing field map calculated from $\Tbf$ and $\Tbf_{inv}$ using Eq. (\ref{invC})\\
\label{alg:InOutFlow}
\end{algorithm}

Algorithm \ref{alg:ComputeT} estimates the motion field between fixed and moving images through the implemented DIR algorithm, which incorporates the suggested local Robin boundary conditions, via Eq. (\ref{eq:PBEv}). It corresponds to the blue blocks delineated in Figure \ref{fig:recap}. Note that $i$ is a local variable determing the iteration number (max=1000).

\begin{algorithm}
\caption{Calculation of the motion field estimate}
\SetKwInOut{Input}{Input}
\SetKwInOut{Output}{Output}
\Input{$(I, J, \Tbf, \textbf{g})$ = (fix image, moving image, motion field, incoming/outgoing flow)}
\Output{$\Tbf$ = motion field}
$i \gets 0$\\
\While{$i < 1000$}{
$D \gets$ data fidelity term calculated from $I$, $J$ and $\Tbf$ using Eq. (\ref{data_fidelity})\\
Update $\Tbf$ boundaries according to $\textbf{g}$ (Appendix \ref{app:BC_AR})\\
$J_{cur} \gets J$ transformed with $\Tbf$\\
$\Tbf \gets \mathcal{L}(I, J_{cur}, D, \Tbf)$ \hspace{170pt} \\
Update $\Tbf$ boundaries according to $\textbf{g}$ (Appendix \ref{app:BC_AR})\\
$i \gets i + 1$\\
}
\label{alg:ComputeT}
\end{algorithm}

\section*{References}
\typeout{}
\bibliographystyle{dcu}
\bibliography{2024_DIR_BC}

\end{document}